\documentclass[preprint,12pt]{elsarticle}

\usepackage{graphicx}
\usepackage{amssymb, xcolor, amsmath, soul, caption}
\usepackage{subfigure}

\usepackage{lineno}

\usepackage{bm}

\DeclareMathOperator*{\argmin}{arg\,min}
\DeclareMathOperator*{\expec}{\mathbb{E}}

\journal{Journal Name}

\begin{document}

\begin{frontmatter}


\title{Bayesian Inference with Generative Adversarial Network Priors}

\author{Dhruv Patel}
\ead{dhruvvpa@usc.edu}
\author{Assad A. Oberai \corref{cor1}}
\ead{aoberai@usc.edu}
\cortext[cor1]{Corresponding author}
\address{Aerospace and Mechanical Engineering, University of Southern California, Los Angeles, CA}




\begin{abstract}
Bayesian inference is used extensively to infer and to quantify the uncertainty in a field of interest from a measurement of a related field when the two are linked by a physical model. Despite its many applications, Bayesian inference faces challenges when inferring fields that have discrete representations of large dimension, and/or have prior distributions that are difficult to represent mathematically. In this manuscript we consider the use of Generative Adversarial Networks (GANs) in addressing these challenges. A GAN is a type of deep neural network equipped with the ability to learn the distribution implied by multiple samples of a given field. Once trained on these samples, the generator component of a GAN maps the iid components of a low-dimensional latent vector to an approximation of the distribution of the field of interest. In this work we demonstrate how this approximate distribution may be used as a prior in a Bayesian update, and how it addresses the challenges associated with characterizing complex prior distributions and the large dimension of the inferred field. We demonstrate the efficacy of this approach by applying it to the problem of inferring and quantifying uncertainty in the initial temperature field in a heat conduction problem from a noisy measurement of the temperature at later time. 
\end{abstract}


\end{frontmatter}


\section{Introduction}
\label{S:1}

Bayesian inference is a well-established technique for quantifying uncertainties in inverse problems that are constrained by physical principles \cite{kaipio2006statistical,dashti2016bayesian, Polpo2018}. It has found applications in diverse fields  such as geophysics \cite{Gouveia1997, Malinverno2002, Martin2012, Isaac2015}, climate modeling \cite{Jackson2004}, chemical kinetics \cite{Najm2009}, heat conduction \cite{Wang_2004}, astrophysics \cite{Loredo1990, AsensioRamos2007},  materials modeling \cite{Sabin2000} and the detection and diagnosis of disease \cite{Siltanen2003, Kolehmainen2006}. The two critical ingredients of a Bayesian inference problem are - an informative prior representing the prior belief about the parameters to be inferred and an efficient method for sampling from the posterior distribution. In this manuscript we describe how certain deep generative techniques can be effectively used in these roles. In this section, we provide a brief introduction to Bayesian inference as it is applied to solving inverse problems and to generative adversarial networks (GANs), which are a popular class of deep generative algorithms. Then in the following sections, we described how GANs may be used within a Bayesian context. 

\subsection{Bayesian inference} 
We consider the setting where we wish to infer a vector of parameters $\bm{x} \in \mathbb{R}^N$ from the measurement of a related vector $\bm{y} \in \mathbb{R}^P$, where the two are related through a forward model $\bm{y} = \bm{f}(\bm{x})$. A noisy measurement of $\bm{y}$ is denoted by $\hat{\bm{y}}$,
\begin{eqnarray}
\hat{\bm{y}} = \bm{f}(\bm{x}) + \bm{\eta}, \label{eq:meas}
\end{eqnarray}
where the vector $\bm{\eta}  \in \mathbb{R}^P$ represents noise. 
While the forward map $\bm{f}$ is typically well-posed, its inverse is not, and hence to infer $\bm{x}$ from the measurement $\hat{\bm{y}}$ requires techniques that account for this ill-posedness. Classical techniques based on regularization tackle this ill-posedness by using additional information about the sought solution field explicitly or implicitly \cite{tarantola2005inverse}. Bayesian inference offers a different approach to this problem by modeling the unknown solution as well as the measurements as random variables. This statistical framework addresses  the ill-posedness of the inverse problem, and allows for the characterization of the uncertainty in the inferred solution.

The notion of a prior distribution plays a key role in Bayesian inference. It is assumed that through multiple observations of the field $\bm{x}$, denoted by the set $\mathcal{S} = \{\bm{x}^{(1)}, \cdots, \bm{x}^{(S)}\}$, we have prior knowledge of $\bm{x}$ that can be utilized when inferring $\bm{x}$ from $\hat{\bm{y}}$. This is used to build, or intuit, a prior distribution for $\bm{x}$, denoted by $p^{\rm prior}_{X}(\bm{x})$. Some typical examples of priors include Gaussian process prior with specified co-variance kernels, Gaussian Markov random fields \cite{Fahrmeir2001}, Gaussian priors defined through differential operators \cite{Stuart2010}, and hierarchical Gaussian priors \cite{Marzouk2009, Calvetti2008}. These priors promote some smoothness or structure in the inferred solution and can be expressed explicitly in an analytical form.

Another key component of Bayesian inference is a distribution that represents the likelihood of $\bm{y}$ given an instance of $\bm{x}$, denoted by $p^{\rm l}(\bm{y}|\bm{x})$. This is often determined by the distribution of the error in the model, denoted by $p_{\eta}$, which captures both model and measurement errors. Given this, and  an additive model for noise (\ref{eq:meas}), the posterior distribution of $\bm{x}$, determined using Bayes' theorem after accounting for the observation $\hat{\bm{y}}$ is given by, 
\begin{eqnarray}
p^{\rm post}_{X}(\bm{x}|\bm{y}) &=& \frac{1}{\mathbb{Z}} p^{\rm l}(\bm{y}|\bm{x}) p^{\rm prior}_{X}(\bm{x}) \nonumber \\
    &=& \frac{1}{\mathbb{Z}} p_\eta(\hat{\bm{y}} - \bm{f}(\bm{x})) p^{\rm prior}_{X}(\bm{x}). \label{eq:bayes1}
\end{eqnarray}
Here,
\begin{eqnarray*}
\mathbb{Z} \equiv \int_{\Omega_x} p^{\rm l}(\bm{y}|\bm{x}) p^{\rm prior}_{X}(\bm{x}) d \bm{x},
\end{eqnarray*}
is the prior-predictive distribution of $\bm{y}$ and ensures that the posterior is a true distribution and integrates to one.

The posterior distribution above completely characterizes the uncertainty in $\bm{x}$; however for vectors of large dimension (that is, large $N$) characterizing this distribution explicitly is a challenging task. Consequently the expression above is used to perform tasks that are more manageable. These include determining estimates such as the maximum a-posteriori estimate (MAP) which represents the value of $\bm{x}$ that maximizes the posterior distribution, expanding the posterior distribution in terms of other distributions that are simpler to work with \cite{Bui-Thanh2012}, or using techniques like Markov Chain Monte-Carlo (MCMC) to generate samples that are ``close'' to the samples generated by the true posterior distribution \cite{han2001markov,parno2018transport}.

Despite its numerous applications and successes in solving inverse problems, Bayesian inference faces significant challenges. These include 
\begin{enumerate}
    \item defining a reliable and informative prior distribution for $\bm{x}$ when the set $\mathcal{S} = \{\bm{x}^{(1)}, \cdots, \bm{x}^{(S)}\}$  is difficult to characterize mathematically. 
    \item efficiently sampling from the posterior distribution when the dimension of $\bm{x}$ ($N$) is large,a typical situation in many practical science and engineering applications.
\end{enumerate}

\subsection{Generative adversarial networks} 
\begin{figure}[htbp] 
   \centering
   \includegraphics[width=4in]{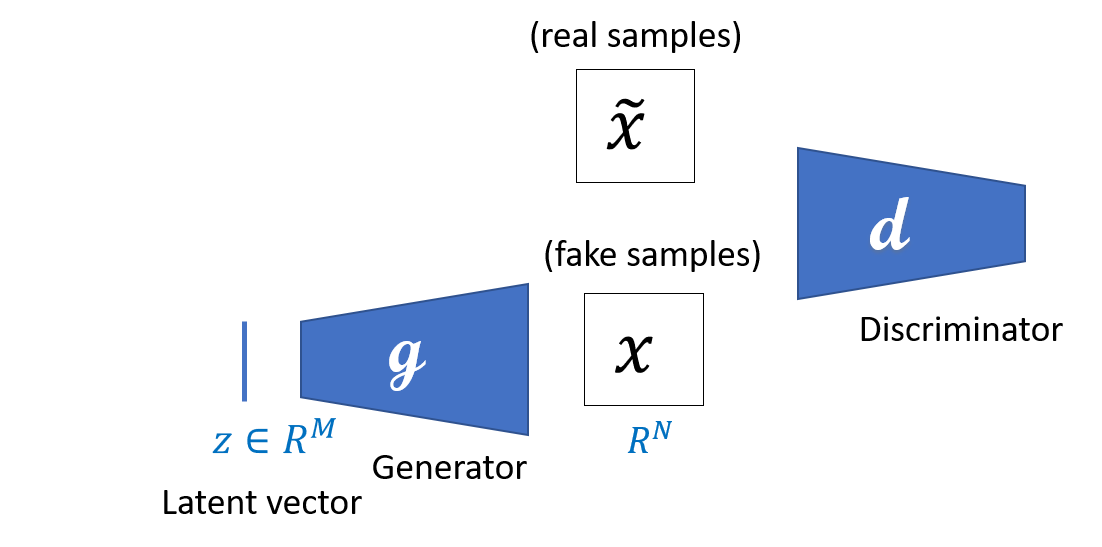} 
   \caption{Schematic diagram of a GAN.}
   \label{fig:gans}
\end{figure}

Generative adversarial networks, or GANs, are a class of generative deep neural networks based on a two-player min-max game that have found many applications since their advent  \cite{goodfellow2014generative}. As shown in Figure \ref{fig:gans}, they comprise of a generator $\bm{g}$ that maps a latent vector $\bm{z} \in \mathbb{R}^M$ to $\bm{x} \in \mathbb{R}^N$, where typically, $M \ll N$. The components of the latent vector are selected from a simple distribution, typically a Gaussian or a uniform distribution. The generator up-scales these components through successive application of non-linear transformation at each layer, whose parameters are learned by the algorithm during training.

The other component of a GAN is a discriminator, which is also composed of successive non-linear transformations. However, these transformations are designed to down-scale the original input. The final few layers of the discriminator are fully connected neural networks which lead to a simple classifier (like a soft-max, for example). The discriminator maps an input field, $\bm{x}$, to a binary variable, which indicates whether the input is ``real'' or ``fake''. The discriminator's weights are also learned during training. 

The generator and the discriminator are trained in an adversarial manner. The training data for the discriminator is comprised of the set of real instances of $\bm{x}$, that is $\mathcal{S}$, and a set of ``fake'' instances generated by the generator, along with the corresponding label: fake or real. The loss function is driven by the accuracy with which the discriminator correctly labels each image. The generator is trained by passing its output through the discriminator and requiring it to be labeled as ``real." Thus while the generator is trained to ``fool'' the discriminator, the discriminator is trained so as not to be fooled by the generator. 

By carefully selecting the loss function the adversarial training process described above can be interpreted as ensuring similarity (in an appropriate measure) between the true distribution of $\bm{x}$, denoted by $p^{\rm true}_X (\bm{x})$, and the distribution of the samples generated by the generator, denoted by $p^{\rm gen}_X (\bm{x})$. A particular family of GANs, called the Wasserstein GAN, which minimizes the Wasserstein metric between $p^{\rm true}_X (\bm{x})$ and $p^{\rm gen}_X (\bm{x})$, has emerged as one of the most popular type of GAN due to its better stability properties \cite{Arjovsky2017, gulrajani2017improved}. 

In several applications, GANs have demonstrated a remarkable ability to approximate the underlying true distribution  \cite{goodfellow2014generative, Makhzani2015, Dumoulin2016,Mescheder2017, Brock2018, Karras2017, fedus2018maskgan,tulyakov2018mocogan, Ma2017, Wang2017}. Further, since samples from the approximate distribution are generated by sampling from the much simpler distribution of the latent vector $\bm{z}$ (whose dimension is much smaller than that of  $\bm{x}$), they have been applied to generate numerous samples of $\bm{x}$ consistent with $p^{\rm true}_X (\bm{x})$.

\subsection{Related work}
The main idea developed in this paper involves training a GAN using the sample set $\mathcal{S}$, and then using the distribution learned by the GAN as the prior distribution in Bayesian inference. This leads to a useful method for representing complex prior distributions and an efficient approach for sampling from the posterior distribution by re-writing it in terms of the latent vector $\bm{z}$.

The solution of an inverse problem using sample-based priors has a rich history (see \cite{Vauhkonen1997,Calvetti2005} for example). As does the idea of reducing the dimension of the parameter space by mapping it to a lower-dimensional space \cite{Marzouk2009,Lieberman2010}. However, the use of GANs in these tasks is novel. 

Recently, a number of authors have considered the use machine learning-based methods for solving inverse problems. These include the use of convolutional neural networks (CNNs) to solve physics-driven inverse problems \cite{Adler2017,Jin2017,patel2019circumventing}, and GANs to solve problems in computer vision \cite{Chang, Kupyn2018, Yang2018, Lediga, Anirudh2018, pix2pix2016, CycleGAN2017,  Kim2017}.  There is also a growing body of work dedicated to using GANs to learn regularizers in solving inverse problems \cite{lunz2018adversarial} and in compressed sensing \cite{bora2017compressed, bora2018ambientgan, kabkab2018task, wu2019deep, shah2018solving}. However, these approaches differs from ours in at least two significant ways. First, they solve the inverse problem as an optimization problem and do not rely on Bayesian inference; as a result, regularization is added in an ad-hoc manner, and no attempt is made to quantify the uncertainty in the inferred field. Second, the forward map is assumed to satisfy an extension of the restricted isometry property, which may not be the case for forward maps induced by physics-based operators. 

More recently, the approach described in \cite{adler2018deep} utilizes GANs in a Bayesian setting; however the GAN is trained to approximate the posterior distribution (and not the prior, as in our case), and training is done in a supervised fashion. That is, paired samples of the measurement $\hat{\bm{y}}$ and the corresponding true solution $\bm{x}$ are required. In contrast, our approach is  unsupervised, where we require only samples of the true solution $\bm{x}$ to train the GAN prior.

We note that deep learning based Bayesian networks are another avenue of exciting research \cite{gal2016uncertainty, Gal2015}. However, these algorithms are significantly different from the approach described in this paper. In these algorithms, conventional artificial neural networks are extended to a regime where network weights are stochastic parameters whose distribution is determined using a Bayesian inference problem. 

The layout of the remainder of this paper is as follows. In Section 2, we develop a formulation for Bayesian inference when the prior distribution is defined by a GAN and describe techniques for sampling from this distribution. Thereafter in Section 3, we utilize these techniques to solve an inverse problem and quantify uncertainty in our solution. We end with conclusions in Section 4.

\section{Problem Formulation}
\label{S:2}
The central idea of this paper is to train a GAN using the sample set $\mathcal{S}$ and then use the distribution defined by the GAN as the prior distribution in Bayesian inference. As described in this section, this leads to a useful method for representing complex prior distributions and an efficient approach for sampling from the posterior.

 Let $\mathcal{S}$ denote the set of instances of vector $\bm{x}$ sampled from the true distribution, $p^{\rm true}_X (\bm{x})$. Further, let $\bm{z} \sim p_Z(\bm{z})$ characterize the latent vector space and $\bm{g}(\bm{z})$ be the generator of a GAN trained using $\mathcal{S}$. Then according to \cite{goodfellow2014generative}, with infinite capacity and sufficient data, the generator learns the true distribution. That is, 
 \begin{eqnarray}
    p^{\rm gen}_X (\bm{x}) = p^{\rm true}_X (\bm{x}). \label{eq:perfectgan}
\end{eqnarray}
The distribution $p^{\rm gen}_X(\bm{x})$ is defined as 
\begin{eqnarray}
\bm{x} \sim p^{\rm gen}_X(\bm{x}) \Rightarrow \bm{x} = \bm{g}(\bm{z}), \bm{z} \sim p_Z(\bm{z}). \label{eq:pgen}
\end{eqnarray}
Here $p_Z$ is the multivariate distribution of the latent vector whose components are iid and typically conform to a Gaussian or a uniform distribution. The equation above implies that the GAN  creates synthetic samples of $\bm{x}$ by first sampling $\bm{z}$ from $p_Z$ and then passing these samples through the generator.

Now consider a measurement $\hat{\bm{y}}$ from which we would like to infer the posterior distribution of $\bm{x}$. For this we use (\ref{eq:bayes1}) and set the prior distribution to be equal to the true distribution, that is $p^{\rm prior}_X = p^{\rm true}_X$. Then, under the conditions of infinite capacity of the GAN, and sufficient data, from (\ref{eq:perfectgan}), this is the same as setting $p^{\rm prior}_X = p^{\rm gen}_X$ in this formula. Therefore,
\begin{eqnarray}
p^{\rm post}_{X}(\bm{x}|\bm{y}) 
    &=& \frac{1}{\mathbb{Z}} p_\eta(\hat{\bm{y}} - \bm{f}(\bm{x})) p^{\rm gen}_{X}(\bm{x}). \label{eq:bayes2}
\end{eqnarray}
Now for any $l(\bm{x})$, we have 
\begin{eqnarray}
\expec_{\bm{x} \sim p^{\rm post}_{X}}[l(\bm{x})] &=&  \frac{1}{\mathbb{Z}} \expec_{\bm{x} \sim p^{\rm gen}_{X}}[l(\bm{x}) p_\eta(\hat{\bm{y}} - \bm{f}(\bm{x})) ], \qquad \qquad \mbox{From (\ref{eq:bayes2})} \nonumber \\
    &=& \frac{1}{\mathbb{Z}} \expec_{\bm{z} \sim p_{Z}}[l(\bm{g}(\bm{z})) p_\eta(\hat{\bm{y}} - \bm{f}(\bm{g}(\bm{z}))) ] , \qquad \mbox{From (\ref{eq:pgen})} \nonumber \\
    &=&  \expec_{\bm{z} \sim p^{\rm post}_{Z}}[l(\bm{g}(\bm{z}))], \label{eq:pxpost1}
\end{eqnarray}
where $\expec$ is the expectation operator, and 
\begin{eqnarray}
p^{\rm post}_Z(\bm{z}|\bm{y}) \equiv \frac{1}{\mathbb{Z}} p_\eta(\hat{\bm{y}} - \bm{f}(\bm{g}(\bm{z}))) p_{Z}(\bm{z}). \label{eq:pzpost}
\end{eqnarray}

Note that the distribution $p^{\rm post}_Z$ is the analog of $p^{\rm post}_X$ in the latent vector space. The measurement $\hat{\bm{y}}$ updates the prior distribution for $\bm{x}$ to the posterior distribution. Similarly, it updates the prior distribution for $\bm{z}$, $p_Z$, to the posterior distribution, $p^{\rm post}_Z$, defined above. 

Equation (\ref{eq:pxpost1}) implies that sampling from the posterior distribution for $\bm{x}$ is equivalent to sampling from the posterior distribution for $\bm{z}$ and transforming the sample through the generator $\bm{g}$. That is,
\begin{eqnarray}
\bm{x} \sim p^{\rm post}_X (\bm{x}| \bm{y}) \Rightarrow \bm{x} = \bm{g}(\bm{z}), \bm{z} \sim p^{\rm post}_Z(\bm{z}|\bm{y}). \label{eq:pxpost}
\end{eqnarray}
Since the dimension of $\bm{z}$ is typically much smaller than that of $\bm{x}$, and since the operation of the generator is typically inexpensive to compute, this represents an efficient approach to sampling from the posterior of $\bm{x}$. 

Note that the left hand side of (\ref{eq:pxpost1}) is the expression for a population parameter of the posterior, defined by $\overline{l(\bm{x})} \equiv \expec_{\bm{x} \sim p^{\rm post}_{X}}[l(\bm{x})]$. The right hand sides of the last two lines of this equation describe how this parameter may be evaluated by sampling $\bm{z}$ (instead of $\bm{x}$) from either $p_Z$ or $p_Z^{\rm post}$. In the following section we describe sums that approximate these integrals. 



\subsection{Sampling from the posterior distribution}
We consider the following scenario:
\begin{itemize}
    \item We wish to infer and characterize the uncertainty in the vector of parameters $\bm{x}$ from a noisy measurement of $\bm{y}$ denoted by $\hat{\bm{y}}$ in  (\ref{eq:meas}), where $\bm{f}$ is a known map that connects $\bm{x}$ and $\bm{y}$. 
    \item We have several prior measurements of plausible $\bm{x}$, contained in the set $\mathcal{S}$.
\end{itemize}

For this problem we propose the following algorithm that accounts for the prior information in $\mathcal{S}$ and the ``new'' measurement $\hat{\bm{y}}$ through a Bayesian update: 
\begin{enumerate}
    \item Train a GAN with a generator $\bm{g}(\bm{z})$ on $\mathcal{S}$.
    \item Sample $\bm{x}$ from $p^{\rm post}_X(\bm{x}| \bm{y})$ given in (\ref{eq:pxpost}). 
\end{enumerate}
With sufficient capacity in the GAN and with sufficient training, the posterior obtained using this algorithm will converge to the true posterior. Further, since GANs can be used to represent complex distributions efficiently, this algorithm provides a means of including complex priors that are solely defined by samples within a Bayesian update. 

As mentioned earlier, an efficient approach to sampling from $p^{\rm post}_X(\bm{x}| \bm{y})$ is to recognize that the dimension of $\bm{z}$ is typically much smaller ($10^1$ - $10^2$) than that of $\bm{x}$ ($10^4$ - $10^7$). We now describe two approaches for estimating population parameters of the posterior that make use of this observation. 

\paragraph{Monte-Carlo (MC) approximation} The first approach is based on a Monte-Carlo approximation of a population parameter of the posterior distribution. This integral, which is defined in the second line of (\ref{eq:pxpost1}), may be approximated as,
\begin{eqnarray}
\overline{l (\bm{x})} \equiv \expec_{\bm{x} \sim p^{\rm post}_{X}}[l(\bm{x})] 
    \approx \frac{\sum_{n = 1}^{N_{\rm samp}} l(\bm{g}(\bm{z})) p_\eta(\hat{\bm{y}} - \bm{f}(\bm{g}(\bm{z}))) }{\sum_{n = 1}^{N_{\rm samp}}  p_\eta(\hat{\bm{y}} - \bm{f}(\bm{g}(\bm{z})))} , \qquad \bm{z} \sim  p_Z(\bm{z}). \label{eq:mc}
\end{eqnarray}
In the equation above, the numerator is obtained from a MC approximation of the integral in (\ref{eq:pxpost1}), and the denominator is obtained from a MC approximation of the scaling parameter $\mathbb{Z}$. We note that the sampling in this approach is rather simple since in a typical GAN the $z_i$s are sampled from simple distributions like a Gaussian or a uniform distribution. 

\paragraph{Markov-Chain Monte-Carlo (MCMC) approximation} In many applications we anticipate that the likelihood will tend to concentrate the distribution of latent vector $\bm{z}$ to a small region within $\Omega_z$. Thus the MC sampling described above may be inefficient since it will include regions where the likelihood will take on very small values. A more efficient approach will be to generate an MCMC approximation $p^{\rm mcmc}_Z(\bm{z}|\bm{y}) \approx p^{\rm post}_Z(\bm{z}| \bm{y})$ using the definition in
(\ref{eq:pzpost}), and thereafter sample $\bm{z}$ from this distribution. Then the corresponding sample for $\bm{x}$ is given by $\bm{x} = \bm{g}(\bm{z})$, and from the third line of (\ref{eq:pxpost1}), any desired population parameter may be approximated as 
\begin{eqnarray}
 \overline{l (\bm{x})}\equiv \expec_{\bm{x} \sim p^{\rm post}_{X}}[l(\bm{x})]  \approx \frac{1}{N_{\rm samp}} \sum_{n = 1}^{N_{\rm samp}} l(\bm{g}(\bm{z})), \qquad \bm{z} \sim  p^{\rm mcmc}_Z(\bm{z}| \bm{y}).  \label{eq:mcmc} 
\end{eqnarray}


\subsection{Expression for the maximum a-posteriori estimate}

The techniques described in the previous section focused on sampling from the posterior distribution and computing approximations to  population parameters. These techniques are general in that they can be applied in conjunction with any distribution used to model noise and the latent space vector; that is, any choice of $p_\eta$ (likelihood) and $p_Z$ (prior). In this section we consider the special case when Gaussian models are used for noise and the latent vector. In this case, we can derive a simple optimization algorithm to determine the maximum a-posteriori estimate (MAP) for $p^{\rm post}_Z(\bm{z}|\bm{y})$ as described below. This point is denoted by $\bm{z}^{\rm map}$ in the latent vector space and represents the most likely value of the latent vector in the posterior distribution. It is likely that the operation of the generator on $\bm{z}^{\rm map}$, that is $\bm{g}(\bm{z}^{\rm map})$, will yield a value that is close to $\bm{x}^{\rm map}$, and may be considered as a likely solution to the inference problem. 

We consider the case when the components of the latent vector are iid with a normal distribution with zero mean and unit variance. This is often the case in many typical applications of GANs. Further, we assume that the components of noise vector are defined by a normal distribution with zero mean and a covariance matrix $\bm{\Sigma}$. Using these assumptions in (\ref{eq:pzpost}), we have 
\begin{eqnarray}
p^{\rm post}_Z(\bm{z}|\bm{y}) \propto \exp{\Big( -\frac{1}{2}\overbrace{\big( | \bm{\Sigma}^{-1/2} (\hat{\bm{y}} - \bm{f}(\bm{g}(\bm{z})))|^2  + |\bm{z}|^2 \big)}^{\equiv r(\bm{z})} \Big)}.
\label{eq:postZ}
\end{eqnarray}
The MAP estimate for this distribution is obtained by maximizing the argument of the exponential. That is 
\begin{eqnarray}
\bm{z}^{\rm map} = \argmin_{\bm{z}} r(\bm{z}). \label{eq:zmap}
\end{eqnarray}
This minimization problem may be solved using any gradient-based optimization algorithm. The input to this algorithm is the gradient of the functional $r$ with respect to $\bm{z}$, which is given by 
\begin{eqnarray}
\frac{\partial r}{\partial \bm{z}} = \bm{H}^T(\bm{z}) \bm{\Sigma}^{-1} (\bm{f}(\bm{g}(\bm{z})) - \hat{\bm{y}}) + \bm{z},  \label{eq:gradpi}
\end{eqnarray}
where the matrix $\bm{H}$ is defined as 
\begin{eqnarray}
\bm{H} \equiv \frac{\partial \bm{f}(\bm{g}(\bm{z}))}{\partial \bm{z}} = \frac{\partial \bm{f}}{\partial \bm{x}} \frac{\partial \bm{g}}{\partial \bm{z}}.  
\end{eqnarray}
Here $\frac{\partial \bm{f}}{\partial \bm{x}}$ is the derivative of the forward map $\bm{f}$ with respect to its input $\bm{x}$, and $\frac{\partial \bm{g}}{\partial \bm{z}}$ is the derivative of the generator output with respect to the latent vector. In evaluating the gradient above we need to evaluate the operation of the matrices $\frac{\partial \bm{f}}{\partial \bm{x}}$ and $\frac{\partial \bm{g}}{\partial \bm{z}}$ on a vector, and not the matrices themselves. The operation of $\frac{\partial \bm{g}}{\partial \bm{z}}$ on a vector can be determined using a back-propagation algorithm with the GAN; while the operation of $\frac{\partial \bm{f}}{\partial \bm{x}}$ can be determined by making use of the adjoint of the linearization of the forward operator. In the case of a linear inverse problem, this matrix is equal to the forward map itself.

Once $\bm{z}^{\rm map}$ is determined, one may evaluate $\bm{g}(\bm{z}^{\rm map})$ by using the GAN generator. This represents the value of the field we wish to infer at the most likely value value of latent vector. Note that this is not the same as the MAP estimate of $p^{\rm post}_X(\bm{x}|\bm{y})$. 

\paragraph{Remark} It is interesting to note that in a typical Bayesian inverse problem (that does not use GANs as priors) under additive Gaussian noise and Gaussian prior with $\bm{\Sigma}_{\rm prior}$ as covariance, the posterior distribution is given by,
\begin{eqnarray}
p^{\rm post}_X(\bm{x}|\bm{y}) \propto \exp{\Big( -\frac{1}{2}\big( | \bm{\Sigma}^{-1/2} (\hat{\bm{y}} - \bm{f}(\bm{x}))|^2  + | {\bm{\Sigma}_{\rm prior}^{-1/2}} \bm{x}|^2 \big) \Big)}.
\label{eq:postx}
\end{eqnarray}
Seeking $\bm{x}^{\rm map}$ leads to an optimization problem that is similar to the one for $\bm{z}^{\rm map}$ (\ref{eq:gradpi}). However, there are two crucial differences. First, it is harder problem to solve since the optimization variable is $\bm{x}$, whose dimension is greater than that of $\bm{z}$. Second, while different choices of  $\bm{\Sigma}_{\rm prior}$ lead to different types of regularizations for the MAP (like $L_2$ or $H^1$), all of these tend to smooth the solution and none allow for the preservation of sharp edges, which is critical in medical imaging and other applications. Total variation (TV) type regularization strategies do allow for these types of solutions; however they do not translate to conjugate priors with Gaussian likelihood when viewed from a Bayesian perspective. In contrast to this, when using a GAN as a prior we can allow for sharp variations in $\bm{x}$, while still enjoying the benefit of conjugate priors for determining $\bm{z}^{\rm map}$. 

\paragraph{Summary} We have described three algorithms for probing the posterior distribution when the prior is defined by a GAN. These include an MC (\ref{eq:mc}) and an MCMC estimate (\ref{eq:mcmc}) of a given population parameter and a MAP estimate that is applicable to additive Gaussian noise with a Gaussian prior for the latent vector (\ref{eq:zmap}). In the following section we apply these algorithms to  a canonical inverse problem.

\section{Numerical results} 

In this section we describe a numerical example where we utilize a GAN (Wasserstein GAN, in particular) as a prior in a Bayesian inference problem. We first train a GAN on a set of realizations of the field to be inferred. Thereafter, using this GAN as a prior, and a single instance of a noisy measurement, we infer the desired field using a Bayesian update.

Since our goal is to validate the approach developed in this paper, we consider examples where we know the ``correct solution.'' In particular, we generate, rather than measure, the input data as follows:
\begin{enumerate}
    \item We select a stochastic parametric representation for the inferred field and generate multiple samples from it. This gives us the set $\mathcal{S}$, which is used to train the GAN prior. 
    \item We sample once more from this parametric representation to generate the ``target'' instance of the inferred field. This is denoted by $\bm{x}^*$.
    \item This field is transformed by the forward operator to generate a noise-free version of the measured field, $\bm{y} = \bm{f}(\bm{x}^*)$. This measurement is corrupted with additive noise drawn from a known distribution to generate the measurement, $\hat{\bm{y}} = \bm{y}^*+ \bm{\eta}$. 
\end{enumerate}   

Once this input data is generated, following the approach described in the previous section we:
\begin{enumerate}
    \item Use a MC approximation to evaluate the MAP ($\bm{x}^{\rm map}$), the mean ($\bar{\bm{x}}$) and the standard deviation (that is the square root of principal diagonal of the auto-covariance of $\bm{x}$) of each component of $\bm{x}$ for the posterior distribution. 
    \item Generate a Markov chain to sample from the posterior an evaluate the statistics listed above. 
    \item Compute the MAP estimate for $\bm{z}$ (denoted by $\bm{z}^{\rm map}$) by solving the minimization problem in (\ref{eq:zmap}) using a gradient-based algorithm, where the gradient is given by (\ref{eq:gradpi}). Then evaluate the corresponding value of $\bm{x}$, given by $\bm{g}(\bm{z}^{\rm map})$.  
\end{enumerate}

The estimates $\bm{x}^{\rm map}$ and $\bm{g}(\bm{z}^{\rm map})$ may be considered our best guess at the correct value of the inferred target field $\bm{x}^*$. Thus the distance between these fields represents the ``error'' introduced in the Bayesian inference. We note that there are three sources of this error: the loss of information inherent in the forward map $\bm{f}$, the noise in the measurement, and the approximations inherent in our algorithm. The approximation errors include the difference between the true prior distribution and the distribution learned by the GAN, and the errors induced by the MC or MCMC sampling. 

\subsection{Inferring the initial state in a heat conduction problem}

\begin{figure}[!ht] 
   \centering
   \includegraphics[width=0.7\linewidth]{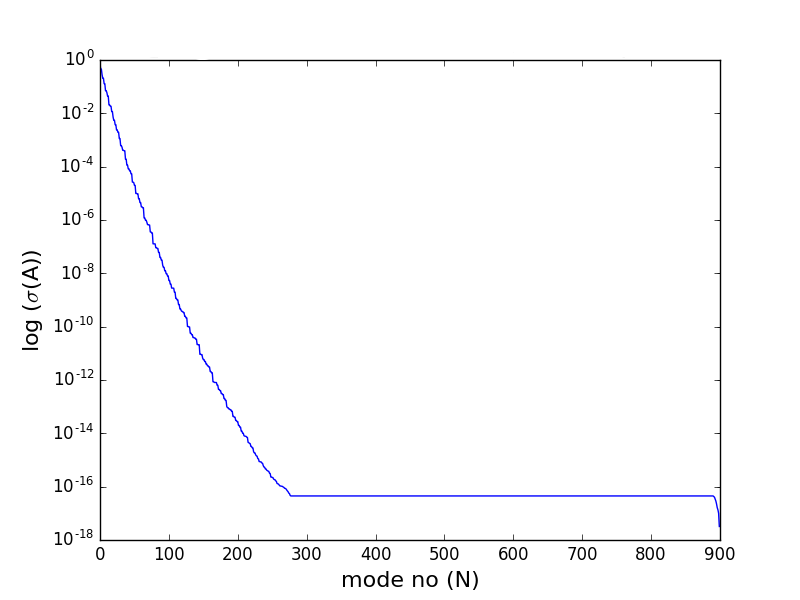} 
   \caption{Eigenvalues of the discretized forward operator $\bm{A}$.}
   \label{fig:sing}
\end{figure}

We apply the Bayesian inference approach with a GAN prior to  the problem of determining the initial temperature distribution of a Fourier solid from a measurement of its current temperature. In this case the field to be inferred ($\bm{x}$) is the initial temperature, which is represented on a $32^2$ grid on a square of edge length $L = 2\pi$ units. The forward operator is defined by the solution of the time-dependent heat conduction problem with uniform conductivity, $\kappa = 0.64$. This operator maps the initial temperature (the quantity to be inferred) to the temperature at time $t = 1$ (the measured quantity, $\bm{y}$). The discrete version of this operator is generated by approximating the time-dependent linear heat conduction equation using central differences in space and backward difference in time. it is given by,
\begin{eqnarray}
\bm{y}  = \bm{A} \bm{x},
\end{eqnarray}
where 
\begin{eqnarray}
\bm{A} = \big( \bm{I} + \Delta t \bm{K}\big)^{-N_t}. 
\end{eqnarray}
In the equation above, $\bm{K}$ is the second-order finite difference approximation of the Laplacian, $\Delta t = 0.01$ is the time step, and $N_t = 100$ is the number of time steps. In Figure \ref{fig:sing}, we have plotted the eigenvalues of this operator as a function of mode number on a log scale. We notice that they decay exponentially with increasing mode number, indicating a significant loss of information in the higher modes. The modes would be used to represent sharp edges and corners in the inferred solution.

\begin{figure}[!ht] 
   \centering
   \includegraphics[width=0.8\linewidth]{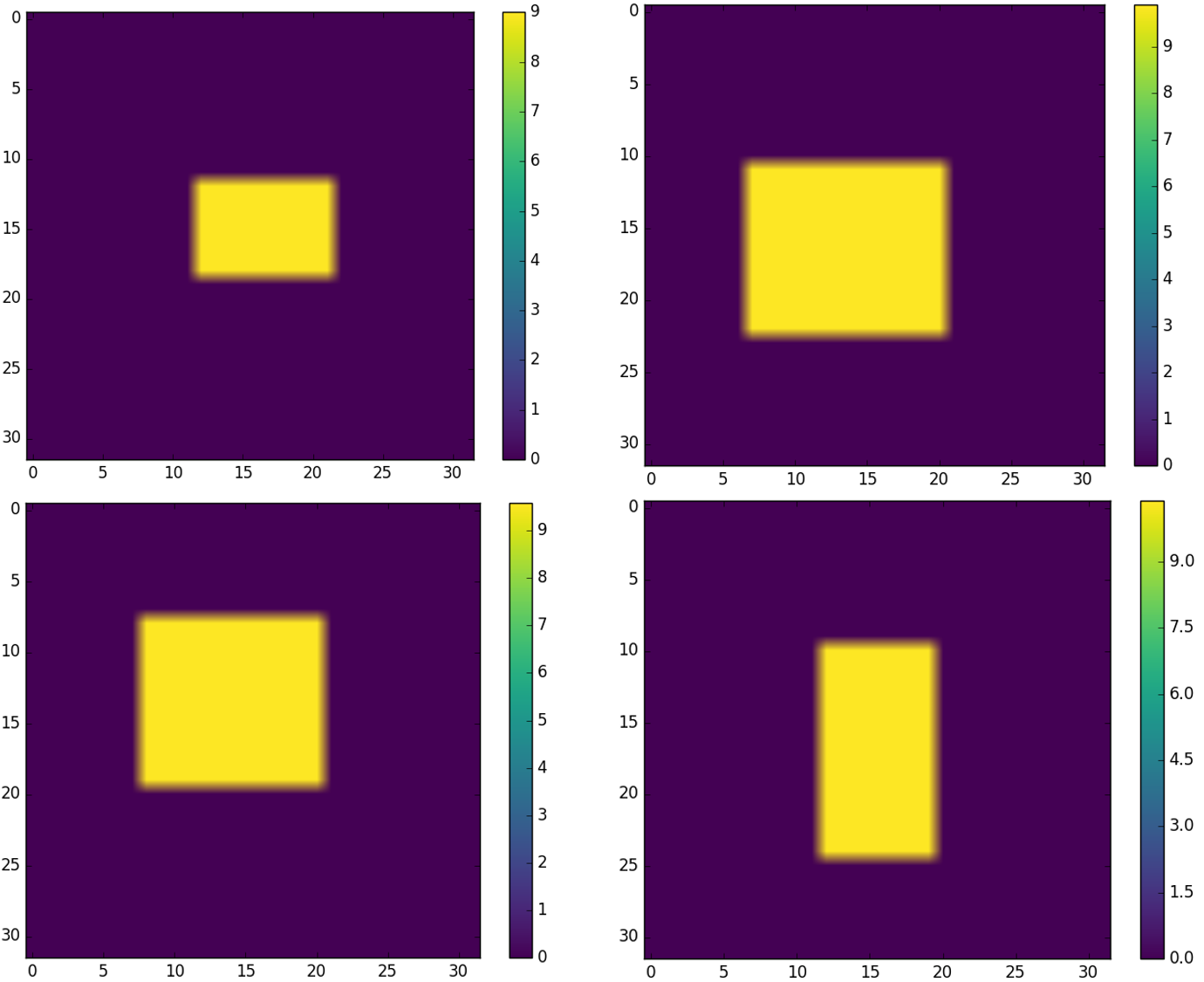} 
   \caption{Sample images from set $\mathcal{S}$ used to train the GAN.}
   \label{fig:real_samples}
\end{figure}

It is assumed that the initial temperature is zero everywhere except in a rectangular region where it is set to a non-zero value. 
The initial temperature field is parameterized by the horizontal and vertical coordinates of the top-left and bottom-right corners of the rectangular region and the value of the temperature field within the rectangular region. Each  parameter is chosen from a uniform distribution. The top-left coordinates are constrained to be in the range $[0.2L, 0.4L]$, while the bottom-right coordinates are constrained to be in the range $[0.6L, 0.8L]$. We note that the vertical axis is positive in the downward direction. The value of temperature inside the rectangular region is a constant constrained to be in the range of $[9, 11]$ units.
Initial temperature fields sampled from this distribution are included in the sample set $\mathcal{S}$ which is used to train the GAN. Four randomly selected samples from this set, which contains 50,000 images, are shown Figure \ref{fig:real_samples}.

We train a Wasserstein GAN (WGAN) with gradient penalty term on the set $\mathcal{S}$ to create a generator to produce synthetic images of the initial temperature field. The detailed architecture of the generator ($\bm{g}$) and the discriminator($\bm{d}$) are shown in Appendix A. The generator consists of 3 residual blocks (see \cite{He2015a}) and 4 convolutional layers and the discriminator consists of 3 residual blocks and 8 convolutional layers. Both the generator and the discriminator were trained using Adam optimizer with equal learning rate of 1e-4 and momentum parameters $\beta_1$ = 0.9 and $\beta_2$ = 0.5. The entire training and inference was performed using Tensorflow \cite{tensorflow2015-whitepaper} on a workstation equipped with dual Nvidia GeForce RTX 2080Ti GPUs.

\begin{figure}[!ht] 
   \centering
   \includegraphics[width=0.8\linewidth]{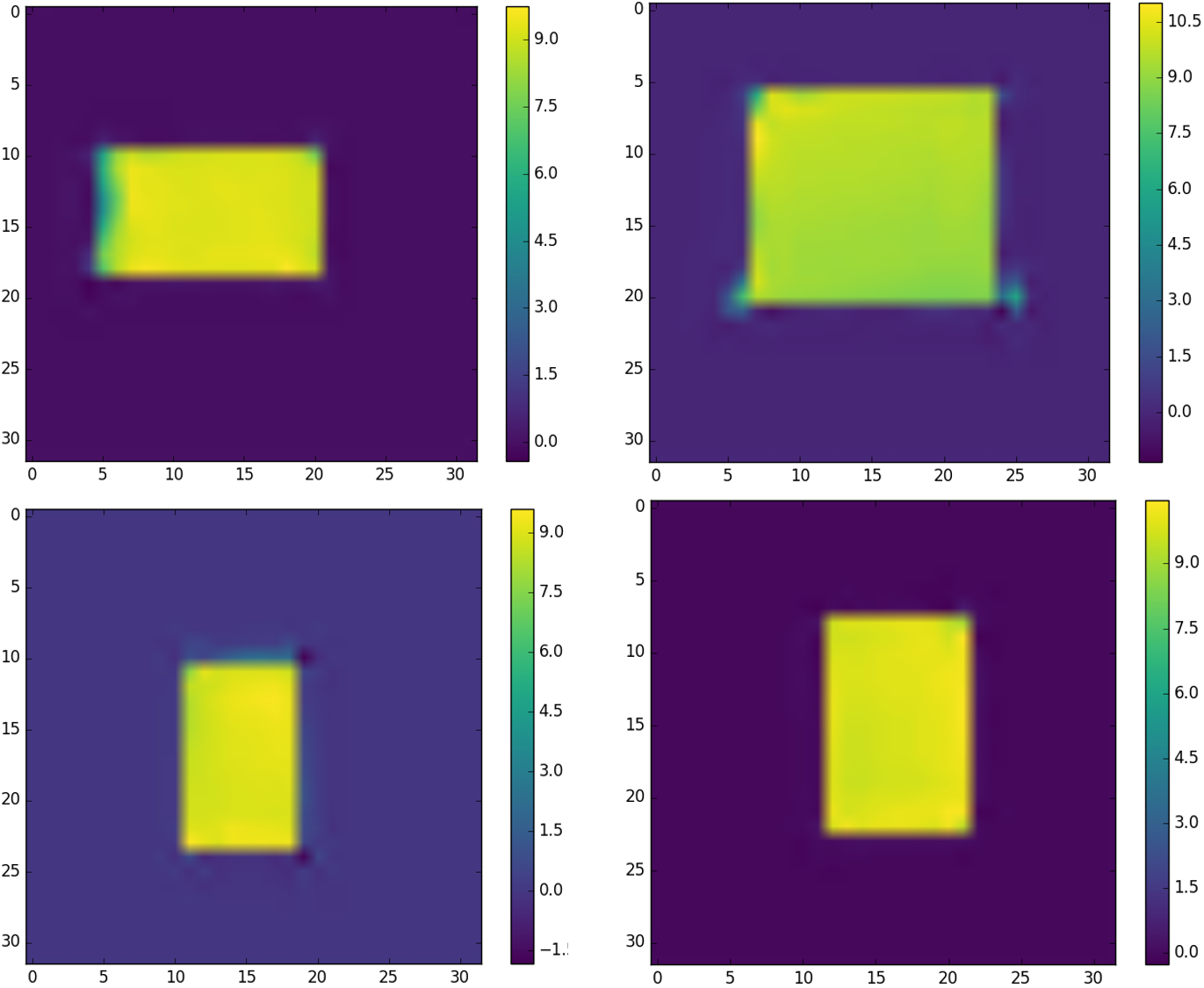} 
   \caption{Sample images produced by trained GAN.}
   \label{fig:fake_samples}
\end{figure}

The latent vector space of the GAN comprises of 8 iid variables conforming to a Gaussian distribution with zero mean and unit variance. The training of the GAN proceeds in the standard adversarial manner (see \cite{gulrajani2017improved}), and the samples generated by it become more realistic as the training progresses. Some representative images produced by the fully-trained GAN are shown in Figure \ref{fig:fake_samples}. From these images we conclude that the GAN is able to replicate the key characteristics of the true data remarkably well. However, we also observe some slight deviations, which could likely be addressed with more training and/or capacity. 

\begin{figure}[!ht]
    \centering
    \subfigure[The target field $\bm{x}^*$.]{\includegraphics[width=0.45\linewidth]{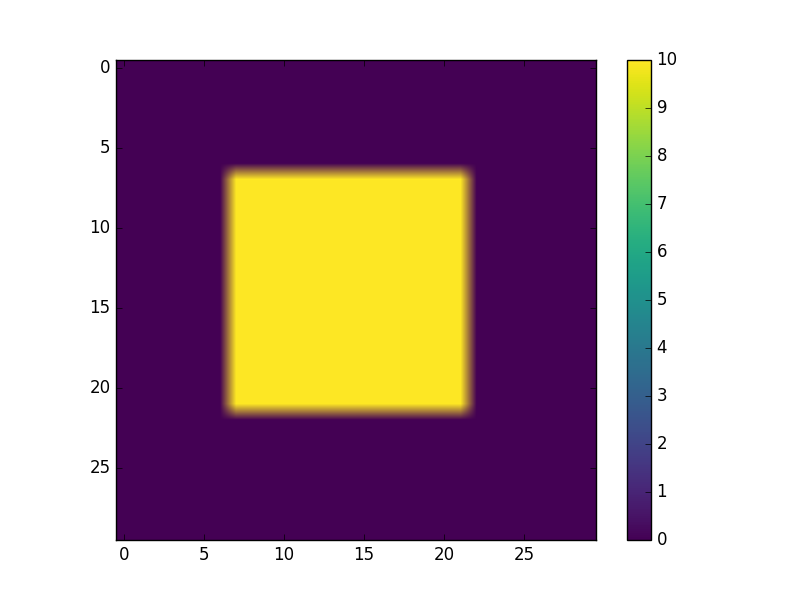}}  
    \subfigure[Measurement without noise $\bm{y}^* = \bm{f}(\bm{x}^*)$.]{\includegraphics[width=0.45\linewidth]{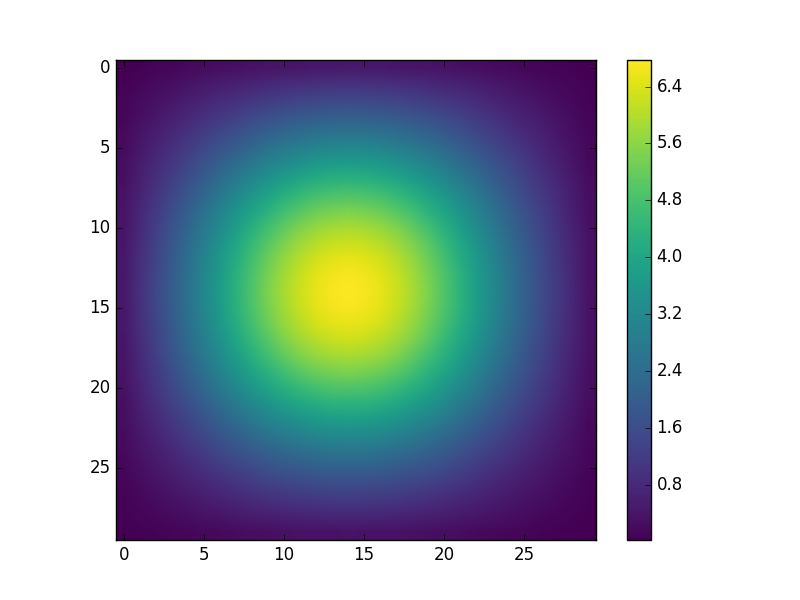}}
    \subfigure[Measurement with noise $\hat{\bm{y}} = \bm{y}^* + \bm{\eta}$.]{\includegraphics[width=0.45\linewidth]{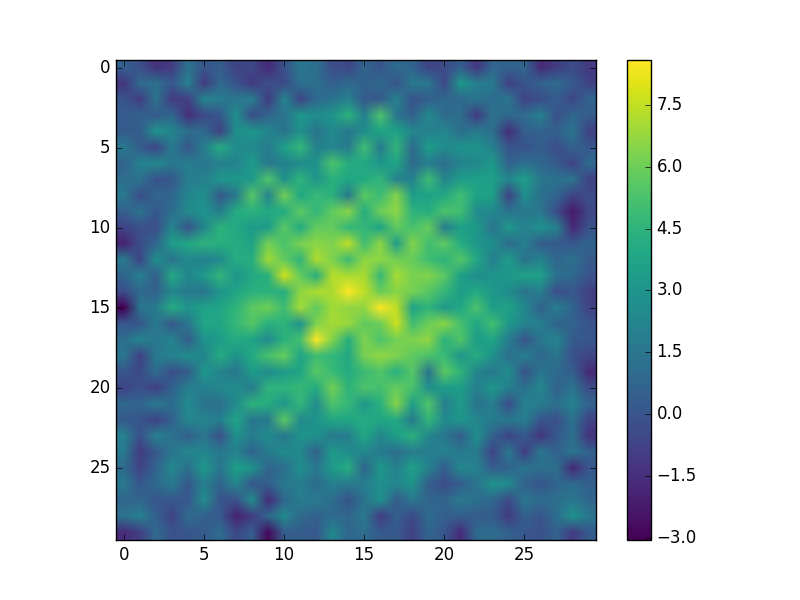}} 
    \caption{The target field and the measurement.} \label{fig:inp1} 
\end{figure}

Next we generate the target field that we wish to infer and the corresponding measurement. As shown in Figure \ref{fig:inp1}a, this comprises of a square patch with edge = $L/2$ centered on center of the total domain. This field is passed through the forward map to generate the noise-free version of the measured field, which is shown in Figure \ref{fig:inp1}b. Thereafter, iid Gaussian noise with zero mean and unit variance is added to this field to generate the synthetic measured field (shown in Figure \ref{fig:inp1}c.). 

Once the generator of the GAN is  trained and the measured field has been computed, we apply the algorithms developed in the previous section to probe the posterior distribution. 

\begin{figure}[!ht]
    \centering
    \subfigure[The target field $\bm{x}^*$.]{\includegraphics[width=0.45\linewidth]{x_true.png}}  
    \subfigure[MAP estimate with a GAN prior $\bm{g}(\bm{z}^{\rm map})$.]{\includegraphics[width=0.45\linewidth]{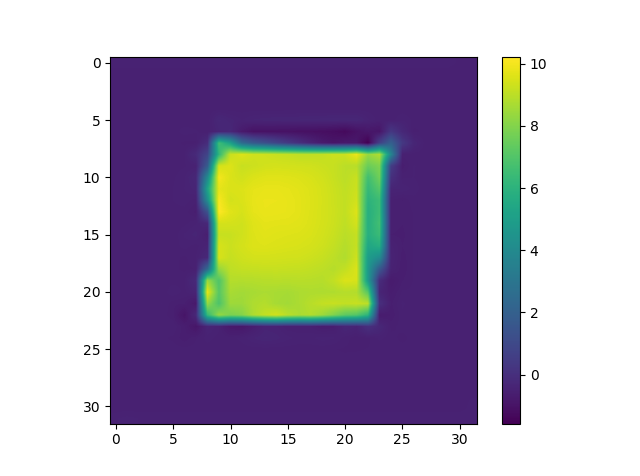}} 
    \subfigure[MAP estimate with an $H^1$ prior.]{\includegraphics[width=0.45\linewidth]{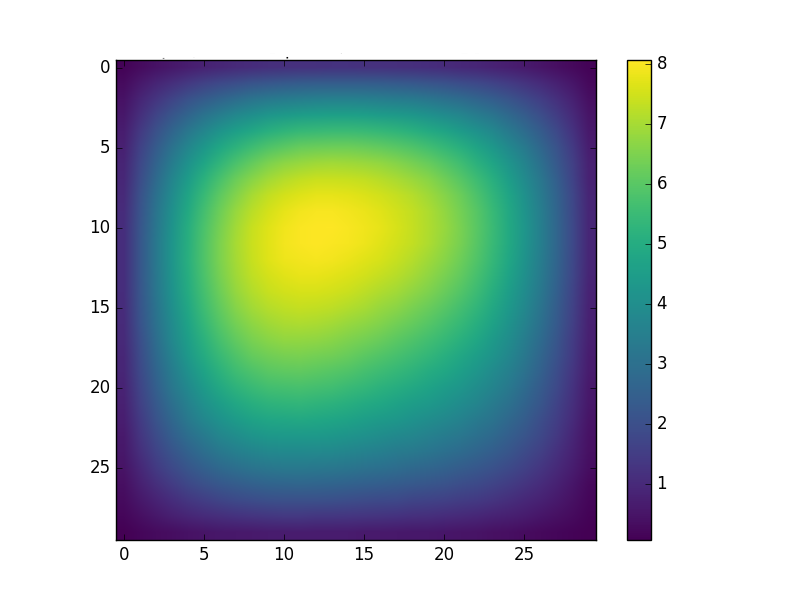}} 
    \subfigure[MAP estimate with an $L_2$ prior.]{\includegraphics[width=0.45\linewidth]{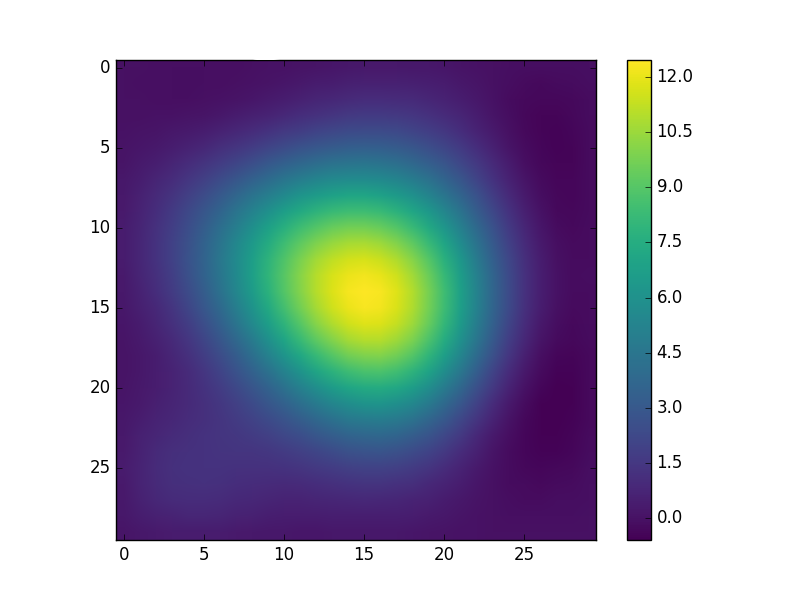}} 
     \subfigure[MAP estimate from MCMC.]{\includegraphics[width=0.45\linewidth]{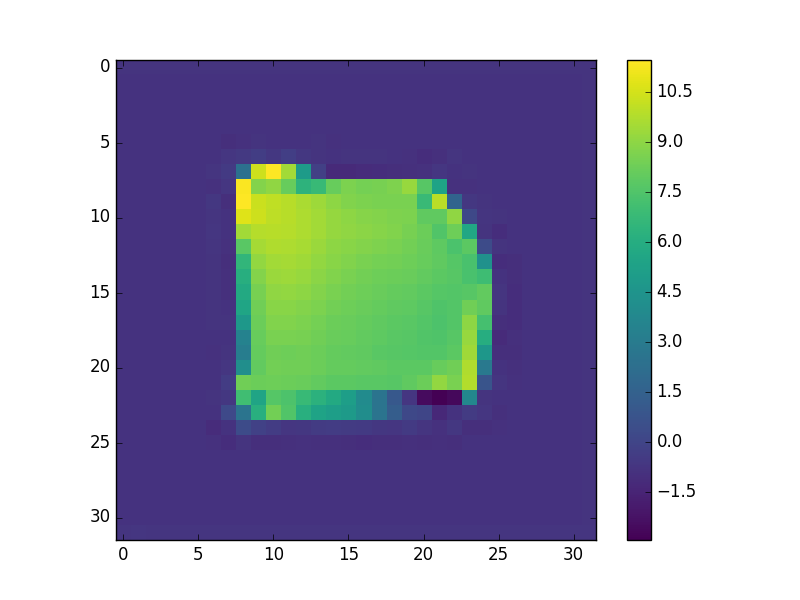}} 
     \subfigure[Point-wise mean from MCMC.]{\includegraphics[width=0.45\linewidth]{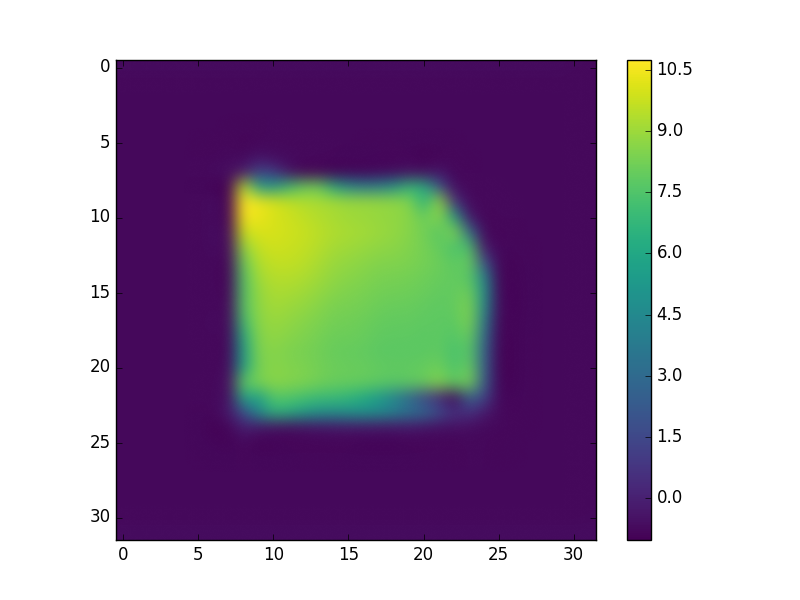}} 
    \caption{Comparison of the true field with the the inferred fields.} \label{fig:maps1} 
\end{figure}

We first use these to determine the MAP estimate for the posterior distribution of the latent vector (denoted by $\bm{z}^{\rm map}$). In order to evaluate this estimate we use a gradient-based algorithm (BFGS) to solve (\ref{eq:zmap}). We use 32 distinct initial guesses for $\bm{z}$, and drive these to values where the gradient, as defined in (\ref{eq:gradpi}), is small. Of these we select the one with the smallest value of $r$ as an approximation to the MAP estimate. The value of $\bm{g}(\bm{z}^{\rm map})$ is shown in Figure \ref{fig:maps1}b. By comparing this with the true value of the inferred field, shown in Figure \ref{fig:maps1}a, we observe that the MAP estimate is very close to the true value. This agreement is even more remarkable if we recognize that the ratio of noise to signal is around $30 \%$, and also compare the MAP estimates obtained using an $L_2$ or an $H^1$ prior (see Figures \ref{fig:maps1}c and d) with the true value. We note that these estimates are very different from the true value, and in both cases the edges and the corners of the initial field are completely lost. In contrast to this, the MAP estimate from the GAN prior is able to retain these features. This is primarily because these characteristics of the spatial distribution of the initial temperature field, that is a rectangular patch with homogeneous temperature distribution, are embedded in the prior.

Next, we consider the results obtained by sampling from the MCMC approximation to the posterior distribution of $\bm{z}$. The MCMC approximation was obtained by applying the random walk Metropolis Hastings \cite{Metropolis1953, Hastings1970} algorithm to the distribution defined in (\ref{eq:pzpost}). The proposal density was Gaussian with zero mean and standard deviation equal to 0.005.

The MCMC approximation to the mean of the inferred field computed using (\ref{eq:mcmc}) is shown in Figure \ref{fig:maps1}f. We observe that the edges and the corners of the temperature field are smeared out. This indicates the uncertainty in recovering the values of the initial field along these locations, which can be attributed to the smoothing nature of the forward operator especially for the higher modes. 

The MAP estimate, $\bm{x}^{\rm map}$, was obtained by selecting the sample with largest posterior density among the MCMC samples, and is shown in Figure \ref{fig:maps1}e. We observe that both in its spatial distribution and quantitative value, it is close to the true distribution. Further, by comparing Figure \ref{fig:maps1}e and \ref{fig:maps1}f, we note that while $\bm{x}^{\rm map}$ and  $\bm{g}(\bm{z}^{\rm map})$ share some common features, they are not identical. This is to be expected because, in general $\bm{x}^{\rm map} \ne \bm{g}(\bm{z}^{\rm map})$. 
 
\begin{figure}[!ht]
    \centering
\includegraphics[width=0.5\linewidth]{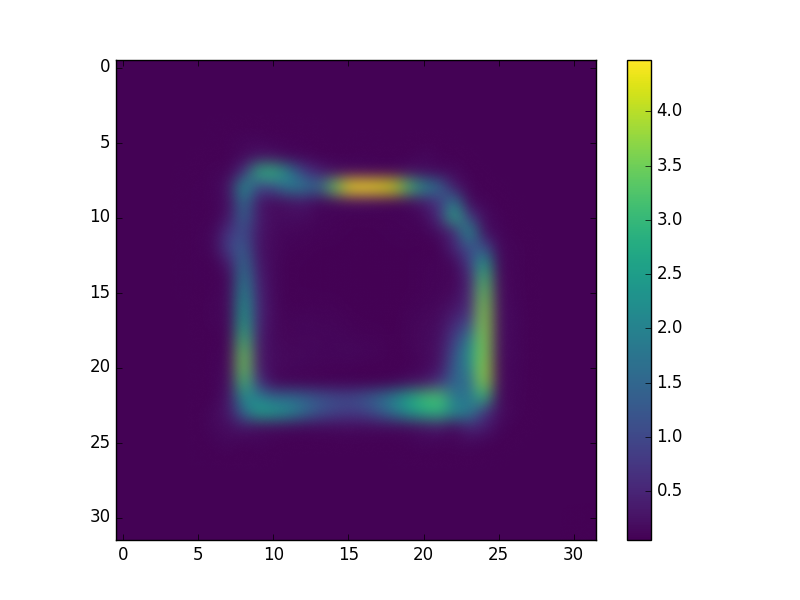}
    \caption{MCMC estimate of point-wise standard deviation.} \label{fig:mcmc} 
\end{figure}

A more precise estimate of the uncertainty in the inferred field can be gleaned by computing the variance of the inferred initial temperature at each spatial location. In Figure \ref{fig:mcmc} we have plotted the point-wise standard deviation (square-root of the diagonal of co-variance) of the inferred field. We observe that the standard deviation is the largest at edges and the corners, where the forward operator has smoothed out the initial data, and thus introduced large levels of uncertainty in the precise location of these features. 

\section{Conclusions}

In this manuscript we have considered the use of the distribution learned by a GAN as a prior in a Bayesian update. We have demonstrated that with sufficient capacity and training of the GAN, the corresponding posterior tends to be the posterior obtained using the true prior density. 

We believe that this approach addresses two challenges that are often encountered in Bayesian inference. First, it facilitates the application of Bayesian inference to cases where the prior is known only through a collection of samples and is difficult to represent through hand-crafted operators. Second, since a typical GAN generates complex distributions for vectors of large dimension by sampling over a much smaller space of latent vectors, it provides an efficient means to sample the posterior distribution. 

In order to make these ideas practical we have proposed two strategies to estimate population parameters for the proposed posterior. The first is a simple Monte-Carlo approximation to the corresponding integrals where the samples are chosen using the prior. The second involves a Markov-Chain Monte-Carlo (MCMC) approximation of the posterior distribution in order to generate the appropriate samples. Further, under the assumptions of Gaussian noise and Gaussian latent vector components, we have described a simple gradient-based approach to recover the maximum a-posteriori (MAP) estimate of the posterior distribution. 

We have demonstrated the utility of these methods on the simple inverse problem of determining the initial temperature field distribution from a noisy measurement of the temperature field at a later time. In this initial test, we have found that the proposed method works well and holds promise for solving more complex problems. 

The work described in this manuscript can be extended along several different avenues. These include (a) applying it to more complex and challenging inverse problems; some with strong nonlinearities, (b) examining the relation between the dimension of the latent vector space and the accuracy of the posterior distribution, (c) the use of other generative machine learning algorithms, such as variation auto-encoders, as priors, and (d) the application of advanced MCMC techniques like a Metropolis-adjusted Langevin algorithm (MALA) \cite{Atchade2006} and Hamiltonian Monte-Carlo methods (HMC) \cite{Hoffman2014, Brooks2012} for accurately and efficiently sampling the posterior distribution.





\bibliographystyle{model1-num-names}
\bibliography{sample.bib}






\newpage
\appendix

\section{Architecture details}
\begin{figure}[!ht]
    \centering
   {\includegraphics[width=0.85\linewidth]{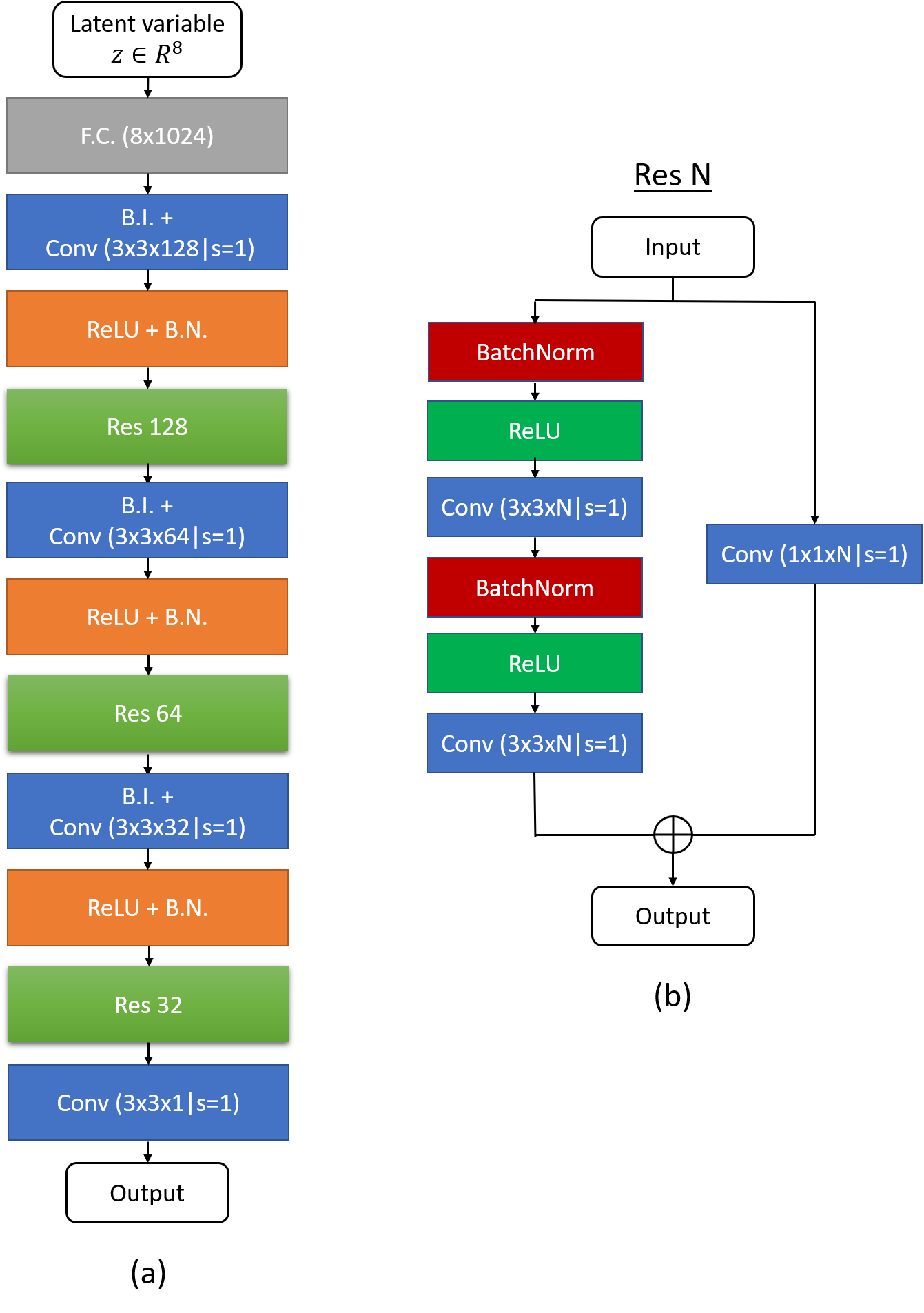}} 
    \caption{(a) Architecture of the generator. (b) Residual block (Res N) used in the generator network. } \label{fig:gen} 
\end{figure}

\begin{figure}[!ht]
    \centering
    {\includegraphics[width=0.85\linewidth]{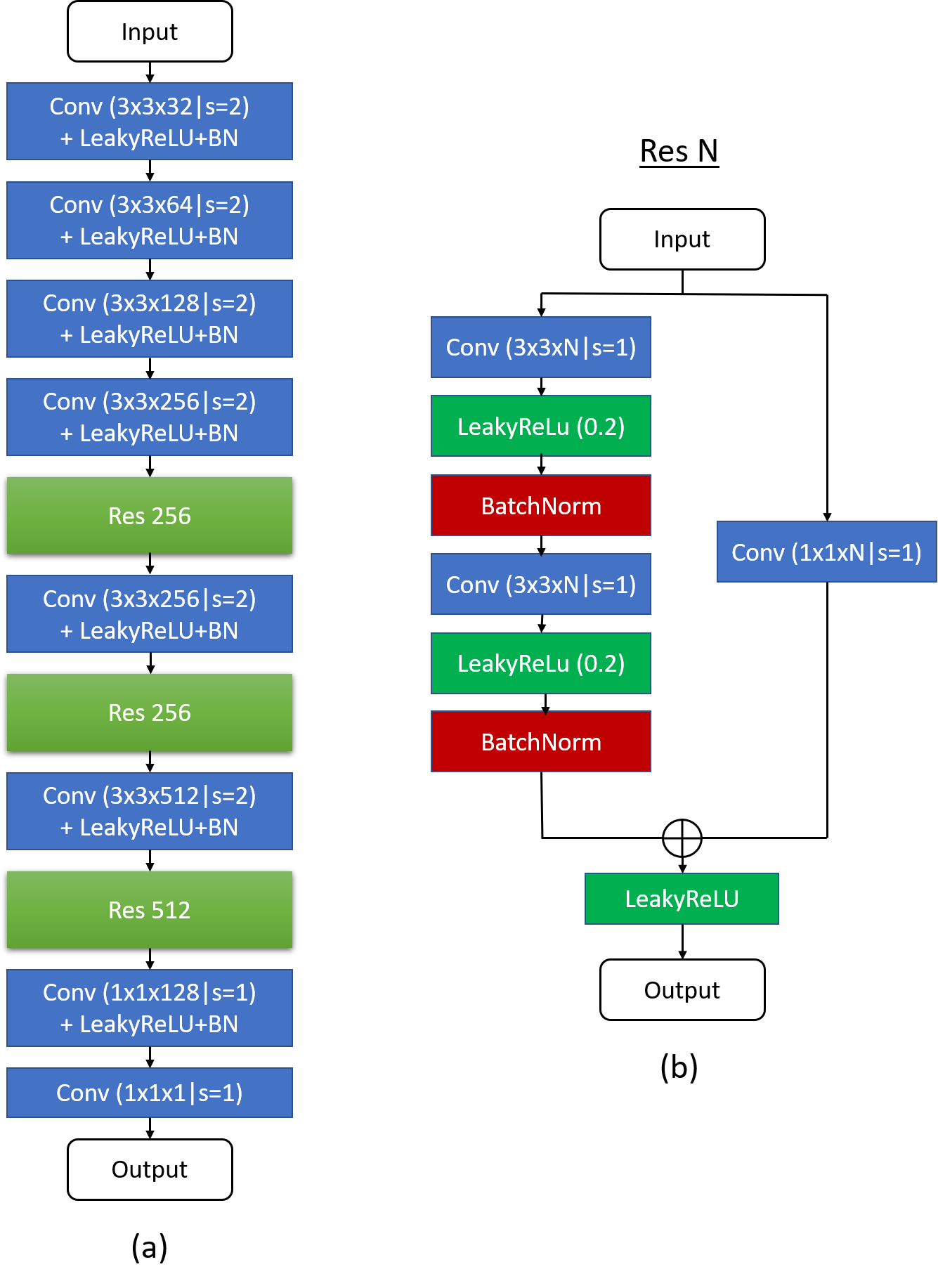}}  
    \caption{(a) Architecture of the discriminator. (b) Residual block (Res N) used in the discriminator network. } \label{fig:disc} 
\end{figure}

The architecture of the generator component of the GAN is shown in Figure \ref{fig:gen}, and the architecture of the discriminator is shown in Figure \ref{fig:disc}. Some notes regarding the nomenclature used in these figures:
\begin{itemize}
    \item Conv($H \times W \times C|=n$) indicates convolution layer with filter size$=H \times W$ and stride $= n$.
    \item BN = Batch normalization
    \item BI = Bilinear interpolation (upscaling by a factor of 2)
    \item Unless otherwise specified all the LeakyReLU activation functions have slope parameter of 0.2
\end{itemize}

\end{document}